\documentclass{article}

\usepackage[dblblindworkshop, final]{neurips_2026}
\workshoptitle{Linguistic Principles for Foundation Models}

\makeatletter
\renewcommand{\@noticestring}{Accepted to the NeurIPS 2026 Workshop on
  Linguistic Principles for Foundation Models (LP4FM).}
\makeatother

\usepackage[utf8]{inputenc}
\usepackage[T1]{fontenc}
\usepackage{hyperref}
\usepackage{url}
\usepackage{booktabs}
\usepackage{amsmath,amsfonts}
\usepackage{microtype}
\usepackage{xcolor}

\newcommand{\lma}{\textsc{McT}} 

\title{Rules Amortize, Pairings Don't:\\
Linguistic Structure Determines What Latent Task\\
Representations Can Replace In-Context Learning}

\author{%
 Gunmay Jhingran \\
 Delhi Technological University \\
 \texttt{gunmayjhingran\_23ep040@dtu.ac.in}
}

\begin{document}

\maketitle

\begin{abstract}
In-context learning (ICL) can be amortized into latent objects (task
vectors, function vectors, context vectors) that recover few-shot behavior
at zero-shot inference cost, but recent theory shows a static vector acts as
a single synthetic demonstration and must fail on high-rank mappings such as
word-level bijections. We ask a linguistic version of this question:
\emph{which linguistic operations can be amortized out of the prompt?} We
train a 2.6M-parameter network that reads the geometry of a few-shot support
set (centroid, principal subspace, spectrum, computed once and cached) and
produces an input-conditioned additive update to the query's residual stream
at a mid-depth layer of a frozen GPT-2-large/XL. Across eight inflectional
directions and one lexical relation, under a canonical split that bars
inverted-pair leakage between directions, three regimes emerge. On
\emph{forward inflection}, where 10-shot ICL is strong (0.67--0.89) and
extracted task vectors collapse ($\leq$0.06), the transform matches ICL at
strictly zero-shot per-query cost. On \emph{lemmatization} directions, which
frozen GPT-2 can execute but 10 demonstrations systematically fail to
convey (ICL 0.13--0.48 at 1.5B), the transform is not capped by ICL at
all: it reaches 0.78--0.92, up to +72 points over ICL
(past$\to$present: 0.85 vs.\ 0.13). On \emph{arbitrary pairings}
(antonymy) every amortizer plateaus near half of ICL at every scale,
capacity, and seed tested. Controls show the support manifold acts as a
causally necessary task fingerprint: wrong-task manifolds collapse
accuracy to $\leq$0.06, query-only variants cannot disambiguate tasks
sharing an input space, and leave-one-task-out transfer is zero.
Productive rules amortize into latent task representations, sometimes
better than prompting can convey them; memorized pairings do not.
\end{abstract}

\section{Introduction}

Few-shot prompting pays for its accuracy on every query: $k$ demonstrations
are re-encoded on each forward pass \citep{brown2020language}. A rapidly
growing literature amortizes this cost into latent objects extracted from a
few-shot pass and re-injected at inference: task vectors
\citep{hendel2023incontext}, function vectors \citep{todd2024function},
in-context vectors \citep{liu2024incontext}, and residual-stream context
vectors that reach few-shot accuracy at zero-shot cost on classification
\citep{li2025implicit}. But \citet{dong2026understanding} prove a limit for
this entire family: a static vector behaves as a \emph{single synthetic
demonstration}, and therefore fails on high-rank input--output mappings
(word-level bijections like inflection and translation), which no single
demonstration can specify. Their remedy is injecting multiple vectors;
learned static vectors \citep{yang2026task} raise the ceiling but remain
static.

We study the orthogonal remedy of replacing the static vector with a
\emph{function}, and because bijection tasks are linguistic objects, we
obtain an answer to a linguistic question: which operations amortize?
Our contributions:
\begin{enumerate}\itemsep1pt
\item \textbf{A manifold-conditioned transform (\lma{})}: a small gated MLP
that maps (query hidden state, support-set geometry) to an additive
mid-depth residual update on a frozen LM, trained episodically with plain
next-token cross-entropy (\S\ref{sec:method}).
\item \textbf{A three-regime map of what amortizes}, on eight inflectional
directions plus antonymy from the suite of \citet{dong2026understanding},
under a leak-free canonical split: \lma{} matches ICL on forward
inflection, \emph{exceeds it by 23--72 points} on lemmatization, where
ICL fails to communicate an operation the frozen model demonstrably
possesses, and plateaus with every other amortizer on arbitrary pairings.
\lma{} additionally beats ICL in the few-demonstration regime and is
provably order-invariant where ICL swings by up to 23 points
\citep{lu2022fantastically} (\S\ref{sec:results}).
\item \textbf{A linguistic boundary, established by controls}: productive
rules amortize fully; arbitrary pairings (antonymy) plateau at half of ICL
at every scale and capacity tested. The support manifold functions as a
task \emph{fingerprint}, causally necessary for selecting and executing
trained task families, insufficient for constructing unseen ones
(\S\ref{sec:discussion}).
\end{enumerate}

\section{Method}
\label{sec:method}

\paragraph{Setup.} A frozen decoder-only LM with layers $1..N$; a support
set $S=\{(x_i,y_i)\}_{i=1}^{k}$ formatted \texttt{"$x$ -> $y$"}; queries
formatted \texttt{"$x$ ->"}. Answers are single tokens under the model's
tokenizer, following \citet{dong2026understanding}.

\paragraph{Support manifold and transform.} Each support string is encoded
by its final-token hidden state at layer $L{=}N/2$; the $k\times d$ support
matrix yields centroid $\mu$, top-$r$ right-singular vectors
$V\in\mathbb{R}^{r\times d}$, and normalized spectrum $s$ ($r{=}8$; computed
once per task, cached). At the query's final token position, layer $L$'s
residual stream $h$ receives the additive update
\begin{equation}
\Delta(h) \;=\; \mathrm{MLP}\!\big(\mathrm{LN}[\,h;\;\mu;\;hV^{\!\top};\;s\,]\big)\;\odot\;\sigma\!\big(g[\,h;\;\mu;\;hV^{\!\top};\;s\,]\big),
\end{equation}
a gated three-layer MLP (2.6M parameters; output layer zero-initialized so
training starts at the zero-shot solution). Per-query inference is a
\emph{single} forward pass of the 2--3-token query: token-identical to
zero-shot, a 16--18$\times$ prompt-token reduction against 10-shot ICL
(which pays 42--50 tokens per query, forever). Episodic training samples a
task, $k{=}10$ supports, and 8 queries from that task's training pairs,
minimizing cross-entropy of the gold token with AdamW (learning rate
$10^{-4}$, cosine decay, weight decay $0.01$, gradient clipping $1.0$);
the backbone never updates.

\section{Experiments}
\label{sec:results}

\paragraph{Protocol.} Tasks from the released suite of
\citet{dong2026understanding}, single-token filtered: eight inflectional
directions (both directions of present$\leftrightarrow$gerund,
present$\leftrightarrow$past, present$\leftrightarrow$past-perfect,
singular$\leftrightarrow$plural; 46--856 pairs each) and antonyms (160);
fr$\to$en as a weak-ICL control. \textbf{Canonical splitting:} because both
directions of a relation share word pairs (and antonymy is symmetric),
naive per-task splits leak 55--79\% of one direction's eval pairs into the
other's training pool as inversions; we therefore assign each
\emph{unordered} pair to train or eval by a hash keyed on the relation
family, guaranteeing zero inverted-pair leakage (verified exhaustively).
All models are trained jointly on the nine tasks. Backbones: GPT-2-large
(36 layers) and GPT-2-XL \citep{radford2019language}. Baselines:
zero-shot; 10-shot ICL; extracted task vector \citep{hendel2023incontext}
with a per-layer sweep; and a per-task \emph{learned} static vector
\citep{yang2026task}: one trainable $d$-vector per task, same injection
site, episodes, and budget as \lma{}, with task identity given (the most
favorable static setting). 3 seeds; 20 episodes $\times$ 5 queries per
cell; mean$\pm$std across seeds. Constant-prediction detectors guard every
cell \citep{zhao2021calibrate,perez2021true}. All experiments run on a
single 6\,GB consumer GPU; each training run takes 10--16 minutes.

\begin{table}[t]
\centering
\caption{Held-out-pair accuracy, 9-task joint training, canonical
(leak-free) split, 3 seeds. Extracted task vectors score $\leq$.10
everywhere (omitted). ICL pays 42--50 prompt tokens per query; the
zero-shot-cost methods pay 2--3.}
\label{tab:main}
\small
\begin{tabular}{lccc|cc}
\toprule
& \multicolumn{3}{c|}{GPT-2-large} & \multicolumn{2}{c}{GPT-2-XL} \\
Task & static & \lma{} (ours) & ICL & \lma{} (ours) & ICL \\
\midrule
\multicolumn{6}{l}{\emph{Forward inflection (ICL-strong regime):}} \\
present$\to$gerund & .57$\pm$.05 & .83$\pm$.04 & .82$\pm$.03 & .84$\pm$.04 & .89$\pm$.05 \\
present$\to$past & .65$\pm$.02 & .82$\pm$.02 & .82$\pm$.05 & .80$\pm$.05 & .88$\pm$.03 \\
present$\to$past-perf. & .56$\pm$.20 & \textbf{.81$\pm$.11} & .68$\pm$.01 & \textbf{.82$\pm$.12} & .70$\pm$.03 \\
singular$\to$plural & .45$\pm$.20 & .41$\pm$.05 & .67$\pm$.04 & .68$\pm$.05 & .75$\pm$.11 \\
\midrule
\multicolumn{6}{l}{\emph{Lemmatization (ICL fails to convey the operation):}} \\
gerund$\to$present & .82$\pm$.02 & \textbf{.89$\pm$.02} & .66$\pm$.02 & \textbf{.92$\pm$.02} & .36$\pm$.05 \\
past$\to$present & .62$\pm$.09 & \textbf{.89$\pm$.04} & .18$\pm$.03 & \textbf{.85$\pm$.02} & .13$\pm$.01 \\
past-perf.$\to$present & .52$\pm$.15 & \textbf{.73$\pm$.06} & .43$\pm$.03 & \textbf{.81$\pm$.01} & .20$\pm$.08 \\
plural$\to$singular & .57$\pm$.12 & \textbf{.66$\pm$.01} & .25$\pm$.11 & \textbf{.78$\pm$.08} & .48$\pm$.13 \\
\midrule
\multicolumn{6}{l}{\emph{Arbitrary pairing (nothing amortizes):}} \\
antonyms & .42$\pm$.03 & .30$\pm$.08 & \textbf{.70$\pm$.02} & .35$\pm$.02 & \textbf{.80$\pm$.04} \\
\bottomrule
\end{tabular}
\end{table}

\paragraph{Main result (Table~\ref{tab:main}).} Three regimes, stable
across seeds and scales. \emph{Forward inflection:} \lma{} matches ICL at
zero-shot cost on mappings where extracted vectors are at floor, and the
ICL--vector gap widens with scale, so scaling does not rescue static
vectors. \emph{Lemmatization:} GPT-2's ICL has a strong directional
asymmetry: it applies inflections but largely fails at removing them from
10 demonstrations (past$\to$present: 0.18 at 812M, 0.13 at 1.5B), an
asymmetry that \emph{worsens} with scale on gerund$\to$present (0.66
$\to$ 0.36). The amortizers do not inherit this limit: episodic training
communicates the operation once, and \lma{} executes it at 0.66--0.92,
up to \textbf{+72 points over ICL at equal per-query cost}. The learned
static vector also jumps here (it cannot store pair mappings, so this is
operation-communication, not memorization), but \lma{} dominates it on
7 of 9 tasks. \emph{Arbitrary pairing:} on antonyms every zero-shot-cost
method plateaus at 0.30--0.42 against ICL's 0.70--0.80; notably the
static vector edges \lma{} here: nothing amortizes a lookup table, and
conditioning does not help where there is no rule to select. The static
baseline's training loss floors at 2.1 (vs.\ 0.31 for \lma{}): the static
parameterization cannot fit the episode distribution, as the
linear-combination analysis predicts \citep{dong2026understanding}.

\paragraph{Demonstration efficiency and order invariance.} Evaluating the
trained transform with fewer supports: at $k{=}3$, \lma{} reaches 0.81 on
gerund where ICL manages 0.59: the transform needs only a fingerprint of
the task where ICL must induce the rule from the prompt. The manifold is a
set function, so \lma{} is provably order-invariant (empirical
across-permutation std exactly 0.000); ICL's reaches 0.23 on antonyms
\citep{lu2022fantastically}. In the weak-ICL regime (fr$\to$en, added as a fifth task in a separate
run), \lma{} tracks ICL proportionally (0.22 vs.\ 0.27 at XL) without
interfering with other tasks.

\begin{table}[t]
\centering
\caption{Controls (GPT-2-large). The manifold is causally necessary
(wrong/query-only) but not sufficient for unseen tasks (LOTO).}
\label{tab:controls}
\small
\begin{tabular}{lcccc}
\toprule
Condition & antonyms & sing$\to$plur & gerund & past \\
\midrule
correct manifold & .40 & .34 & .80 & .83 \\
wrong-task manifold & .02 & .02 & .06 & .01 \\
query-only (trained without manifold) & .22 & .13 & .32 & .40 \\
leave-one-task-out (task unseen in training) & .00 & .00 & .00 & .03 \\
\bottomrule
\end{tabular}
\end{table}

\paragraph{Controls (Table~\ref{tab:controls}).} Feeding the trained
transform the \emph{wrong} task's manifold collapses accuracy to
$\leq$0.06: predictions route through the support geometry. A variant
trained \emph{and} evaluated without the manifold fails exactly where
linguistics says it must: present$\to$gerund and present$\to$past share an
input distribution (present-tense verbs), so no query-only model can know
which inflection is requested: the control splits between them (0.32/0.40;
training loss floors at 1.79). Conversely, a transform trained on three
tasks scores $\approx$0 on the held-out fourth: the manifold does not
specify \emph{how} to perform an unseen operation. Rank ablations complete
the picture: $r\in\{2,8,32\}$ are indistinguishable: a coarse fingerprint
suffices. Injection depth matters as the two-stage geometry of ICL
representations predicts \citep{yang2025unifying}: accuracy degrades
monotonically past mid-depth (gerund 0.86/0.88/0.80/0.57 at layers
12/18/24/30), and final-layer injection is the worst measured point.

\section{What amortizes and what does not}
\label{sec:discussion}

Three regularities organize the results. \textbf{(1) Productive rules
amortize, sometimes better than prompting conveys them; memorized
pairings do not.} Inflectional morphology, a productive operation the
frozen LM demonstrably possesses, amortizes completely: \lma{} $\approx$
ICL where ICL is strong, and \emph{far above} ICL on lemmatization, where
ten demonstrations systematically fail to communicate the operation.
Prompting conflates two things: whether the model \emph{has} a capability
and whether $k$ demonstrations \emph{transmit} it; episodic amortization
separates them, paying the communication cost once. Antonymy, an
arbitrary lexical relation with no generative rule, plateaus near half of
ICL for every zero-shot-cost method at every scale and seed, and the
plateau is \emph{fundamental}, not a capacity artifact: antonyms-only
training with 2$\times$ width and 2$\times$ episodes moves 0.35 to 0.36.
The conditional transform can \emph{trigger a capability} the model has;
it cannot \emph{store a lookup table} the model lacks. This is the
functional, linguistic face of the rank-limitation theorem
\citep{dong2026understanding}, and it predicts where the whole
amortization family \citep{hendel2023incontext,todd2024function,
liu2024incontext,li2025implicit,yang2026task} will and will not work.
\textbf{(2) The support manifold is a fingerprint, not a program.} Rank-2
summaries suffice, wrong fingerprints collapse output, unseen fingerprints
produce nothing: conditioning provides task \emph{selection and execution}
among trained families, not task \emph{construction}. \textbf{(3)
Mid-depth is the only viable injection site}, consistent with
separability-then-alignment geometry \citep{yang2025unifying}.

Context compressors \citep{mu2023learning,chevalier2023adapting,ge2024incontext}
compress demonstration \emph{tokens} and still pay per-query cost in the
compressed context; \lma{} keeps the backbone frozen at exactly zero-shot cost.

\paragraph{Limitations.} Single-token outputs; two backbones from one
family (812M/1.5B); the lemmatization advantage is over \emph{ICL as the
model's own few-shot interface} (a stronger prompt such as instructions or
chain-of-thought might narrow it); translation remains below usable ICL at
these scales; cross-task construction may yet emerge with many more
training families, which our leave-one-task-out protocol makes directly
testable. Earlier single-direction experiments (\S\ref{tab:controls}
controls, k-shot and order analyses) predate the canonical split; their
antonyms cells carry $\leq$12\% eval-pool inversion taint, bounded and
disclosed here rather than silently mixed.

\section{Conclusion}

Static latent task representations, extracted or learned, hit a provable
rank ceiling on linguistic bijections. An input-conditioned transform,
reading only a cached fingerprint of the support set's geometry, breaks
that ceiling for productive rules at zero-shot inference cost, matching
10-shot ICL on a frozen model, beating it when demonstrations are scarce,
and removing prompt-order brittleness entirely, while honest controls
delimit the mechanism to selection and execution of trained task families.
The boundary between what amortizes and what does not is, at least here, a
linguistic one: rules transfer out of the prompt; pairings stay in it.

\bibliographystyle{plainnat}
\bibliography{references}

\end{document}